\documentclass[11pt]{article}

\usepackage[final]{acl}

\usepackage{times}
\usepackage{latexsym}
\usepackage{svg}
\usepackage{float}

\usepackage[T1]{fontenc}

\usepackage[utf8]{inputenc}

\usepackage{microtype}

\usepackage{inconsolata}

\usepackage{graphicx}
\usepackage{amsmath}
\usepackage{booktabs}
\usepackage{makecell}    
\usepackage{amssymb}     
\usepackage{pifont}      
\usepackage{cleveref}
\usepackage{multirow}
\usepackage{svg}
\usepackage{CJKutf8} 

\newcommand{\xmark}{\ding{55}} 
\newcommand{\zh}[1]{\begin{CJK}{UTF8}{gkai}#1\end{CJK}}

\title{Introducing Code-Switched Contexts to Cognitively-Inspired Bilingual Model Training}

\author{
    \bf{Zhuojing Huang} ~~~~
    Luise Pohlmann ~~~~
    Lisa Beinborn \\
    University of G\"{o}ttingen, Germany \\ 
    \texttt{firstname.lastname@uni-goettingen.de} \\}

\begin{document}
\maketitle
\begin{abstract}
During language acquisition, bilingual children are regularly exposed to code-switched input and use it as a cognitive scaffold to accelerate vocabulary growth and cross-linguistic syntactic mapping. In contrast, computational bilingual models are conventionally pretrained on interleaved monolingual corpora. While introducing synthetic code-switching during pretraining has become a promising strategy to enhance cross-lingual alignment and downstream performance, the structural and developmental parameters governing the success remain poorly understood. In this work, we investigate the efficiency of training with synthetic code-switched data across two typologically distinct language pairs by controlling two key variables: the structural location of code-switches and the dynamic switching rate across training stages. Our results show that training with code-switched data improves cross-lingual alignment for typologically close languages.\footnote{Code available at \raisebox{-0.33em}{\includegraphics[height=1.2em]{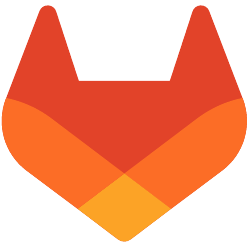}} 
  \href{https://gitlab.gwdg.de/huds/projects/blm2026-code-switching}{GitLab repository}}
\end{abstract}

\section{Introduction}
In many bilingual families, children are regularly exposed to code-switched input during language acquisition \citep{byers-heinlein-2013-parental, BAIL_MORINI_NEWMAN_2015}. Psycholinguistic findings suggest that child code-switching is not a result of confusion or failure \citep{Meisel_2019}. Instead, it is seen as an attempt to communicate given limited linguistic resources \citep{ Paradis_Nicoladis_Genesee_2000, WERKER2008144}. By using code-switching as a scaffold for currently weaker language \citep{YOW_TAN_FLYNN_2018, schneider_morini_dejesus_iglesias}, bilingual children actively draw on their linguistic resources from both languages to support lexical growth \citep{deuchar1994language, Meisel_2019,gaskins-etal-2022-usagebased} with the effect that they need fewer repetitions for novel word acquisition \citep{kaushanskaya_does_2023}. The frequency of language switches varies among bilingual children and over time, as it correlates with language proficiency \citep{kapantzoglou_esparza}. With more stable proficiency in both languages,  code-switching loses its role as a scaffold, and the \emph{switch rate} decreases \citep{smolak_code-switching_2020, kapantzoglou_esparza}.

In contrast to random lexical substitution, the position in an utterance at which code-switching can occur is strongly constrained by structural principles. The \emph{switch point} can be dictated by the structural frame of one of the languages \citep{myers-scotton_duelling_1993} or restricted to positions at which the word order of both languages aligns \citep{poplack_sometimes_1980}. Bilingual children are sensitive to the code-switching patterns in the mostly spoken input they receive from their interlocutors \citep{comeau-etal-2003-modeling}, and they may implicitly infer the syntactic constraints of both linguistic systems from these patterns of switching \citep{CHOUINARD_CLARK_2003}.

Although code-switched input seems beneficial for bilingual proficiency, the training data for bilingual language models usually consists of parallel corpora \citep[e.g.] []{kim-etal-2019-learning, lin-etal-2025-recipe} or interleaved monolingual datasets \citep[e.g.][]{conneau-etal-2020-unsupervised, zheng-etal-2021-low}. While this strategy is effective, it also provides an implicit signal about the language identity of each training example, resulting in cleaner language boundaries than those encountered by bilingual children. To enforce cross-lingual alignment, existing approaches typically include an explicit alignment signal by using a contrastive learning objective \citep{shen-etal-2025-unaligned} or synthetic word-level alignments \citep{kautsar2025paralleltokenizersrethinkingvocabulary}. These approaches primarily operate at the lexical level, failing to capture how children implicitly acquire bilingual lexicons and syntactic structure during code-switching. 

Recent results indicate that code-switched data can make bilingual learning more efficient, improving cross-lingual representation alignment beyond standard parallel corpora \citep{wang-etal-2025-investigating-scaling}, while also benefiting downstream instruction tuning \citep{asano2026bilingualtransfermultilingualcodeswitching} and reducing the need for massive monolingual pretraining corpora \citep{zeng-etal-2026-bringing}. Importantly, code-switched data should not simply be a consequence of mixing languages: the direction and proportion of token insertions can substantially affect performance \citep{wang-etal-2025-investigating-scaling}, while progressively reducing code-switched content toward monolingual input has also been shown to improve cross-language accuracy and alignment \citep{yoo-etal-2025-codeswitching}. 

Despite the promising results, existing work has only examined different factors of code-switched training separately, but not yet how the \emph{switch point} and the \emph{switch rate} jointly shape bilingual representation learning. Therefore, we hypothesize that structurally and developmentally valid code-switched input can facilitate the emergence of shared, syntax-aware bilingual representations without the need for explicit alignment. To test this hypothesis, we train four different code-switching variants across two language pairs with distinct typological distances: English–Dutch \texttt{(eng–nld)} and English–Chinese \texttt{(eng–zho)}, carefully controlling both the \emph{switch rate} and the \emph{switch point}. 

To evaluate our models under developmentally plausible constraints, we use the BabyLM training data for the multilingual track \citep{jumelet-etal-2026-babybabellm} and the 2026 evaluation pipeline \citep{choshen2026babylmturns4goes}, assessing the models' performance on linguistic knowledge and downstream tasks. We show that training on code-switched data generally improves cross-lingual alignment and downstream performance for \texttt{eng–nld} models. For \texttt{eng–zho} models, however, code-switched contexts negatively influence performance in Chinese. To further understand how training with code-switched data affects the learned representation, we examine the geometry of the embedding space and find that code-switching models lead to better cross-lingual alignments for \texttt{eng–nld} models.

\section{Related Work}
While human bilinguals naturally acquire shared semantic concepts through mixed-language input, current bilingual modeling often ignores this aspect. The following sections review the psycholinguistic principles governing human code-switching and present how these insights are currently being integrated into language modeling.

\subsection{Code-Switching in Bilingual Children}
During the language acquisition process, exposure to mixed-language input is a common feature of many bilingual environments \citep{genesee-2008-bilingual, BAIL_MORINI_NEWMAN_2015, kremin-etal-2019-investigating, huang-2025-bilingual}. Despite this input variability, studies show that code-switching carries minimal processing cost in natural settings \citep{blanco-elorrieta-pylkkanen-2017-bilingual, adamou-shen-2019-there}, allowing bilingual children to achieve key linguistic milestones within the same overall time frame as monolingual peers \citep{hambly-etal-2013-influence}. Crucially, \citet{bilson-etal-2015-semantic} suggest that learning a word in one language supports its acquisition in the other, presumably because the underlying semantic representation has already been established, allowing bilingual learners to map an additional lexical form onto an existing concept. As proficiency in both languages stabilizes, code-switching gradually loses its role as an essential scaffolding mechanism, leading to a corresponding increase in the monolingual utterance \citep{smolak_code-switching_2020, kapantzoglou_esparza}. Beyond lexical learning, code-switching during language acquisition is also governed by syntactic constraints. \citet{myers-scotton_duelling_1993} suggests that one ``matrix'' language constructs the morphosyntactic frame, restricting switches predominantly to open-class content words. \citet{poplack_sometimes_1980}, on the other hand, argues that switches occur only at points where the surface word orders of both languages coincide, which opens up more switch points between closely related languages sharing similar syntactic structures.

\subsection{Code-Switching in Bilingual Models}
In language modeling, introducing code-switched contexts enhances cross-lingual representation alignment beyond standard parallel corpora \citep{wang-etal-2025-investigating-scaling}, yielding benefits that extend to instruction-tuning \citep{asano2026bilingualtransfermultilingualcodeswitching} and enabling models to achieve strong performance without relying on massive monolingual pretraining corpora \citep{zeng-etal-2026-bringing}. Code-switching has also been explored as a means of mitigating catastrophic forgetting in multilingual adaptation. By combining POS-based code-switching with a replay adapter strategy, \citet{k-etal-2026-continual} shows that training with code-switched data can help preserve previously acquired knowledge while adapting large language models to low-resource languages. However, these advantages are heavily mediated by factors such as data volume, language-pair similarity, and structural switching dynamics. For instance, \citet{zeng-etal-2026-bringing} show that performance scales logarithmically with the volume of available monolingual data in English–Spanish models. Furthermore, structural mechanics such as insertion direction are critical. For example, in Chinese–English models, increasing Chinese token insertions into English yields continuous performance gains, whereas inserting English tokens into Chinese text is beneficial only up to a specific insertion percentage \citep{wang-etal-2025-investigating-scaling}. To further examine these factors, \citet{yoo-etal-2025-codeswitching} mirror the developmental reduction of human code-switching over time, showing that progressing from frequent intra- and inter-sentential switching to purely monolingual data improves cross-language accuracy and alignment. 

Building on these insights, we compare intra-sentential code-switching strategies within a cognitively-inspired setup for bilingual models. Using strict data ceilings as established in the BabyLM challenge \citep{choshen2026babylmturns4goes}, we evaluate whether different code-switching variants provide consistent benefits across both typologically close (\texttt{eng--nld}) and typologically distant (\texttt{eng--zho}) language pairs.

\section{Methodology}
We train and evaluate one baseline and four code-switching variants for each language pair. Code-switching variants use the same architecture as the baseline, varying only \emph{switch rate} and \emph{switch point}. All models are trained across three random seeds. 

\subsection{Baseline}
Our baseline models are pretrained on a balanced bilingual corpus constructed from the BabyLM 2026 multilingual data collections \citep{jumelet-etal-2026-babybabellm, choshen2026babylmturns4goes}, summing up to 100M tokens.\footnote{Following the BabyLM 2026 guidelines, the Dutch and Chinese sub-corpora are scaled by byte premium factors of $1.05$ and $0.93$, respectively \citep{arnett-etal-2024-bit}.} We adopt the GPT-2-small architecture optimized with a standard causal language modeling objective. A Byte-Pair Encoding (BPE) tokenizer with a vocabulary size of $16,384$ is trained on the unmodified bilingual corpus. We train each model for 10 epochs, with a total exposure of $\sim$1B tokens. Checkpoints are saved at logarithmic intervals, namely, 1M, 10M, 100M, up to 1B words.

\subsection{Code-Switching Models}
\label{sec:method_cs}
We train four code-switching variants that differ in \emph{switch rate} and \emph{switch point} as shown in \Cref{tab:training_paradigms}. These models are optimized on the same training data as the baseline, but the languages are mixed according to the respective switching condition.\footnote{To match the BabyLM challenge guidelines, the word budgets for the code-switching variants are reduced by the number of tokens that are used for creating the bilingual dictionary and POS tagger.}

\begin{table}[h]
\centering
\small
\begin{tabular}{lll}
\toprule
\textbf{Model} &
\textbf{Switch Rate} &
\textbf{Switch Point} \\
\midrule
Baseline            & \phantom{aa}\xmark & \phantom{aaa}\xmark \\
\midrule
Static              & Fixed & Random \\
Curriculum          & Gradual & Random \\
POS-Static         & Fixed & POS-constrained \\
POS-Curriculum      & Gradual & POS-constrained \\
\bottomrule

\end{tabular}
\caption{Overview of code-switching variants.}
\label{tab:training_paradigms}
\end{table}

\paragraph{Switch Rate.} The \emph{switch rate} controls the proportion of words that are switched into another language. While Static models maintain a uniform global token density, i.e., targeting ~10\% switches overall, Curriculum models deliver this same aggregate density through a four-phase dynamic curriculum. Inspired by the correlation between increasing language proficiency and declining code-switching rates in bilingual children \citep{smolak_code-switching_2020, kapantzoglou_esparza}, the curriculum progressively reduces the data proportion for code-switching across four equal word-budget phases, i.e., 90\% $\rightarrow$ 70\% $\rightarrow$ 40\% $\rightarrow$ 0\% of the data are eligible from phase 1 $\rightarrow$ 4. To match the overall 10\% target density, tokens within the active switching windows are sampled at a higher rate (\textasciitilde20\%). Documents are reshuffled within each phase to ensure variable document exposure.

\paragraph{Switch Point.} The \emph{switch point} controls which words are eligible for substitution. Rather than relying on unconstrained random substitution, the POS constrained models only switch at certain word classes. Our setup targets open-class words: Nouns, Verbs, and Adjectives \citep{urbaneja-2020-tu, azuma-meier-1997-open, prasad-rao-2011-code}, motivated by switching constraints observed in human bilingual utterances \citep{myers-scotton_duelling_1993} and findings in language models \citep{k-etal-2026-continual}. This setting aims to minimize grammatical disruption. 

\paragraph{Bilingual Dictionary.} To switch to the equivalent word in the target language, we need bilingual dictionaries for \texttt{eng--nld} and \texttt{eng--zho}. Due to the token constraint of the BabyLM challenge \citep{choshen2026babylmturns4goes}, we do not use pretrained models for creating the dictionaries. Instead, we extract word alignments from parallel subtitle corpora \cite{lison-tiedemann-2016-opensubtitles2016} using \texttt{MGIZA} \citep{gao-vogel-2008-parallel}, a multi-threaded implementation of \texttt{GIZA++} \citep{och03:asc}. Additionally, we train custom POS taggers on Universal Dependencies treebanks \citep{nivre-etal-2017-universal}\footnote{Specifically, we use English Web
Treebank (EWT), Dutch LassySmall, and Chinese PUD treebanks
\citep{silveira-etal-2014-gold,bouma-van-noord-2017-increasing,mcdonald-etal-2013-universal}.} using a BiLSTM architecture without pretrained embeddings. We use the bilingual dictionaries and POS taggers for controlling the \emph{switch rate} and \emph{switch point} of models trained with code-switched data. 

\section{BabyLM Challenge}
We evaluate the models using the benchmark of the BabyLM Multilingual Challenge \citep{choshen2026babylmturns4goes}, which includes tasks for English (\texttt{eng}), Dutch (\texttt{nld}), and Chinese (\texttt{zho}). 

\subsection{Benchmark Tasks} 
We examine both zero-shot linguistic knowledge and fine-tuned downstream performance.\footnote{Detailed tasks lists are provided in \Cref{sec:task_list}.}
\paragraph{Zero-shot.} Zero-shot evaluation measures capabilities acquired during pretraining without additional updates, focusing on grammatical competence, commonsense reasoning, discourse understanding, and semantic composition across languages. These tasks test whether models develop transferable linguistic knowledge, including sensitivity to syntactic structure, understanding of plausible relations, coherent discourse interpretation, and cross-lingual meaning representation.

\paragraph{Fine-tuning.} Fine-tuning evaluation measures the transferability of learned representations by adapting models to downstream tasks. The tasks evaluate broader language understanding abilities, including reasoning, reading comprehension, factual knowledge retrieval, natural language inference, truthfulness, and multilingual classification. 

\subsection{Model Performance}
In this section, we report the model performance change of the four code-switching variants compared to bilingual baselines on the BabyLM benchmark for \texttt{eng--nld} and \texttt{eng--zho}. The results shown in \Cref{tab:cs_delta} are the macro-average of all the tasks, and we report the mean across three random seeds.\footnote{Results for each task are presented in \Cref{app:detailed_results}.}

\begin{table}[t]
\centering
\small
\setlength{\tabcolsep}{3.5pt}
\begin{tabular}{llrr|rr}
\toprule
 & & \multicolumn{2}{c|}{\textbf{eng--nld}} &
     \multicolumn{2}{c}{\textbf{eng--zho}} \\
\textbf{Eval.} & \textbf{Model}
 & $\Delta$eng & $\Delta$nld
 & $\Delta$eng & $\Delta$zho \\
\midrule
ZS & Static
    & -0.08 & \textbf{+0.19}
    & -0.52 & -4.14 \\
   & Curr.
    & -0.32 & -0.19
    & -0.20 & -3.99 \\
   & POS-Static
    & \textbf{+0.38} & -0.30
    & -0.59 & -3.66 \\
   & POS-Curr.
    & -0.54 & -0.46
    & -0.68 & -3.79 \\
\midrule
FT & Static
    & +2.11 & \textbf{+5.23}
    & +6.09 & \textbf{+0.58} \\
   & Curr.
    & +1.44 & +1.63
    & \textbf{+7.05} & -1.06 \\
   & POS-Static
    & -0.84 & +3.31
    & +3.12 & -1.03 \\
   & POS-Curr.
    & \textbf{+2.37} & +4.74
    & +5.73 & -0.23 \\
\bottomrule
\end{tabular}
\caption{Macro-averaged performance changes relative to the bilingual baseline ($\Delta$ in percentage points). The results are averaged across three random seeds. ZS: zero-shot; FT: fine-tuning. Best improvements within each language pair and evaluation setting are in \textbf{bold}.}
\label{tab:cs_delta}
\end{table}

\paragraph{English and Dutch.}
For the \texttt{eng--nld} models, zero-shot performance remains largely unchanged across all code-switching variants, indicating that training with code-switched data preserves general multilingual capabilities. Interestingly, the Static and POS-Static show slightly better overall grammatical knowledge compared to their Curriculum counterparts, indicating that too much code-switching at early training could be slightly harmful for the model's grammatical knowledge. For the fine-tuning tasks, we see that introducing code-switched contexts improves downstream performance for both English and Dutch, with the Static variant achieving the strongest overall results. Overall, these results indicate that training with code-switched data enhances downstream transfer without sacrificing general linguistic knowledge for this closely related language pair.

\paragraph{English and Chinese.}
For the \texttt{eng--zho} models, the result is more asymmetric. For the zero-shot tasks, we see that training with code-switched data is harmful for the model's linguistic knowledge, with the negative effect being particularly pronounced for Chinese. All code-switching variants outperform the bilingual baseline after fine-tuning, but the improvements are driven primarily by the English side, whereas Chinese downstream performance usually remains under the baseline. These results suggest that, for this typologically distant language pair, intra-sentence code-switching affects some aspects of Chinese knowledge acquired during pretraining.

\subsection{Learning Curve}
\begin{figure}[t]
    \centering
        \includegraphics[width=0.95\linewidth]{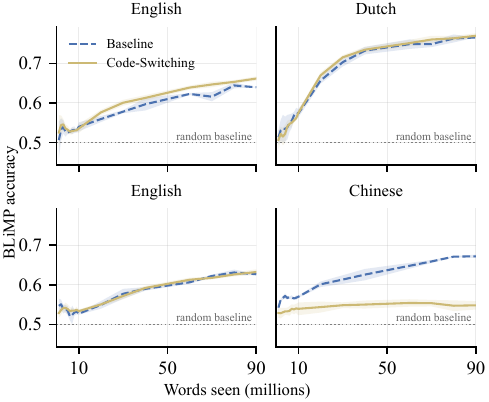}
    \caption{Early learning curve of grammatical knowledge for English–Dutch (top row) and English–Chinese (bottom row) bilingual models. Left column: English BLiMP; right column: target-language BLiMP (Dutch BLiMP for eng--nld, Chinese BLiMP for eng--zho). Each panel shows BLiMP accuracy as a function of words seen during training. The code-switching line represents the mean across four variants. Shaded bands denote standard deviation across random seeds. }
    \label{fig:blimp_curve}
\end{figure}

To inspect learning dynamics, we track learning trajectories on the BLiMP benchmark during early pretraining in \Cref{fig:blimp_curve}. For \texttt{eng--nld}, code-switch models show slightly faster initial progress compared to the baseline. In contrast, \texttt{eng--zho} presents a mixed picture. While English performance remains steady and largely maintains its level relative to the baseline, Chinese accuracy is negatively affected from an early stage. Rather than improving over time, Chinese performance lags behind the baseline early on and fails to recover as pretraining progresses. Overall, these trajectories show that code-switching provides an effective early learning signal for \texttt{eng--nld}, but it has an immediate negative impact on the Chinese knowledge in \texttt{eng--zho}.

\section{Representational Geometry} 
To better understand how training with code-switched data reshapes the internal structure, we examine our models using a range of metrics comparing their global structure, relative alignment across languages, and the effect on cross-lingual transfer.

\subsection{Global Structure} 
As a first sanity check, we measure how introducing code-switched contexts changes the representational structure of the embedding layer compared to a bilingual baseline. 

\begin{figure}[t]
    \centering
        \includegraphics[width=0.95\linewidth]{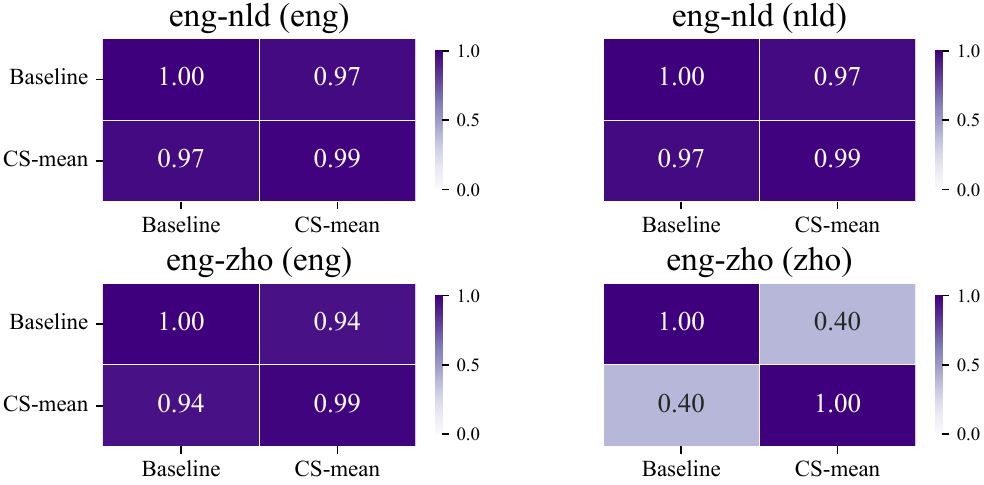}
    \caption{Similarity of the global structure of sentence representations within a language in bilingual models measured by centered kernel alignment (CKA). We compare the \textit{Baseline} model against the mean of code-switching variants (\textit{CS-mean})  for the English--Dutch (\texttt{eng}--\texttt{nld}) models in the top row and the English--Chinese (\texttt{eng}--\texttt{zho}) models in the bottom row. Darker cell colors indicate higher representational similarity.}
    \label{fig:geometry_cka}
\end{figure}

\paragraph{Centered Kernel Alignment.} We measure the similarity of sentence representations within each language using linear Centered Kernel Alignment (CKA) \citep{kornblith2019similarity}. Let $X \in \mathbb{R}^{n \times d_x}$ denote the representation of $n$ sentences in the baseline model and $Y \in \mathbb{R}^{n \times d_y}$ denote the representations of the same sentences in a code-switched model. Linear CKA is defined as
\[
\mathrm{CKA}(X,Y)
=
\frac{\|Y^\top X\|_\mathrm{F}^2}
{\|X^\top X\|_\mathrm{F} \, \|Y^\top Y\|_\mathrm{F}},
\]
where $\|\cdot\|_\mathrm{F}$ denotes the Frobenius norm. CKA quantifies the global similarity of the two representational spaces and is invariant to isotropic scaling and orthogonal transformations. Sentences are represented by the average of their token embeddings. 

\paragraph{Results.}
All code-switching variants lead to highly similar representational structure. We therefore focus on the comparison to the baseline in \Cref{fig:geometry_cka} and average across code-switching conditions and seeds. For the pair \texttt{eng--nld}, the global representational structure remains highly similar for both languages. In contrast, the analysis of the \texttt{eng--zho} models reveals a highly asymmetric pattern. While the structure of the English representations remains relatively close to the baseline, training with code-switched data fundamentally changes the structure of the Chinese sentence representations. Despite this strong shift relative to the baseline, the structure of code-switched variants is highly similar, indicating that the overall effect is driven primarily by the introduction of code-switched context rather than by the specific training variant.

\subsection{Cross-Lingual Alignment} 
After assessing what code-switching changes relative to the baseline, we now examine its effect on the alignment of the two languages within a model. 
\paragraph{Cosine Similarity of Translations.} We measure the alignment between the representation an English input $\mathbf{x}_i$ and the representation of its translation in the target language $\mathbf{y}_i$ by calculating cosine similarity:  
\[
\mathrm{cos}(\mathbf{x}_i,\mathbf{y}_i)
=
\frac{\mathbf{x}_i^\top \mathbf{y}_i}
{\|\mathbf{x}_i\|_2\|\mathbf{y}_i\|_2}.
\]

In addition to the parallel sentence pairs, we perform the same analysis on a curated set of word pairs. We use the translation pairs described in \Cref{sec:method_cs} and additionally examine a list of known cognates with similar form and meaning in English and Dutch \citep{lefever-etal-2020-identifying}.

\begin{figure*}[th]
    \centering
        \includegraphics[width=0.95\linewidth]{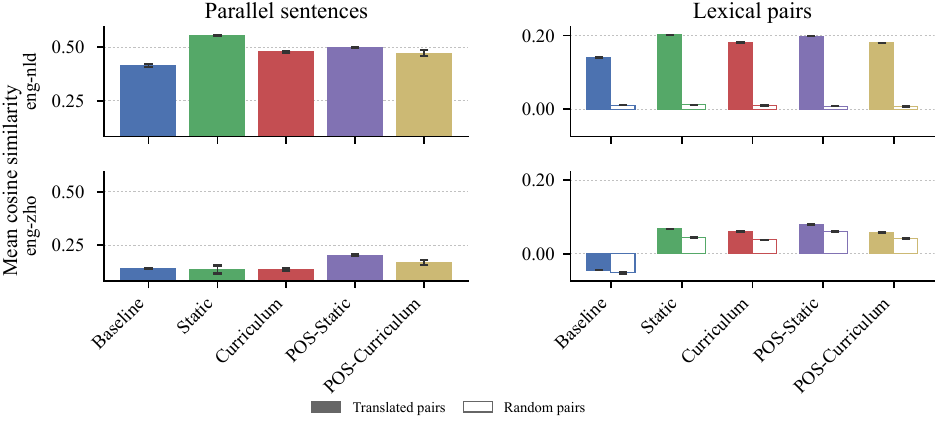}
    \caption{Cross-lingual embedding alignment across models for English--Dutch (eng--nld; top row) and English--Chinese (eng--zho; bottom row). The left column displays the mean cosine similarity for aligned parallel sentences, while the right column shows lexical alignment for curated translation pairs (filled bars) versus randomly matched word pairs (outlined bars).}
    \label{fig:alignment}
\end{figure*}

\begin{figure}[th]
    \centering
    \includegraphics[width=0.95\linewidth]{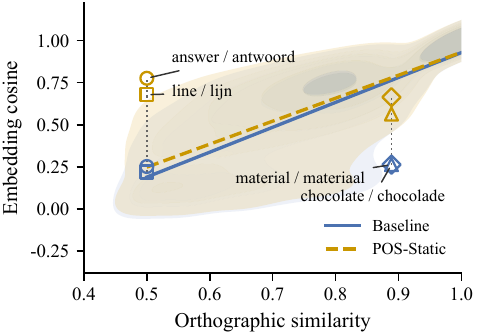}
    \caption{Relationship between orthographic similarity and word embedding cosine similarity for English--Dutch cognates ($n = 5,285$ pairs). Contours indicate kernel density estimates. Dotted connectors link the same cognate pair between the \textit{Baseline} and \textit{POS-Static} setups, illustrating shifts in embedding similarity at equivalent levels of orthographic overlap (e.g., \textit{answer/antwoord}, \textit{line/lijn}), where ideal cross-lingual representations maintain high cosine similarity regardless of distance. Pearson correlations are $r = 0.75$ for Baseline and $r = 0.72$ for POS-Static.}
    \label{fig:cognates}
\end{figure}

\paragraph{Results.} We calculate the average alignment across all input elements and report the delta over the corresponding baseline model in \Cref{fig:alignment}. A higher score indicates that semantically equivalent inputs from the two languages are embedded closer together in the shared representation space. We observe a strong increase in alignment for the \texttt{eng--nld} models both on the sentence level (top left) and the lexical level (top right). Code-switching with a static \emph{switch rate} produces the largest increase, but the differences between code-switching variants remain small. We visualize the lexical alignment for 5,285 \texttt{eng--nld} cognate pairs when training with the POS-Static variant in \Cref{fig:cognates} and compare it to the baseline.\footnote{\Cref{fig:cognates_all_models} in \Cref{app:cognates_alignment} presents plots for all variants.
} It shows that both approaches closely align cognates with high orthographical overlap (\textit{material-materiaal}, \textit{chocolate-chocolate}), but training with code-switched data increases the alignment for more diverging pairs (\textit{answer-antwoord}, \textit{line-lijn}). This indicates that the code-switched context alone can facilitate better cross-lingual semantic alignment beyond surface patterns without introducing an explicit alignment objective during training.

The effect of code-switched training on cross-lingual sentence alignment is generally weaker for the language pair English--Chinese. Only POS-sensitive models slightly increase the sentence-level alignment relative to the baseline. On the lexical level, the difference in cosine similarity between the baseline and the code-switching variants is even more pronounced than for English--Dutch pairs. However, similarity increases for both translation pairs and random pairs, indicating that the changes in representational structure do not necessarily lead to better bilingual understanding.

\subsection{Relative Structure}
We abstract from specific changes in lexical cross-lingual alignment by conducting a second-order representational similarity analysis. 
\paragraph{Representational Similarity Analysis.} 
We calculate the Spearman correlation over the cosine similarities between English representations $X^{(A)}$ and target language representations $X^{(B)}$ for translation pairs. 
While CKA compares feature spaces via linear transformations, RSA evaluates whether two models preserve similar relative distances across lexical translation pairs.

\begin{figure}[th]
    \centering
        \includegraphics[width=0.95\linewidth]{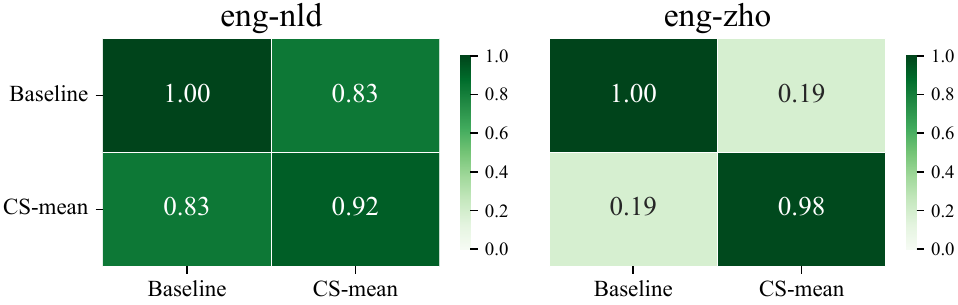}
    \caption{Lexical-level Representational Similarity Analysis (RSA) heatmaps for English---Dutch (\texttt{eng}--\texttt{nld}) and English--Chinese (\texttt{eng}--\texttt{zho}). Panels compare cross-model representational similarity between the \textit{Baseline} model and the average of code-switching variants (\textit{CS-mean}) over paired-token embeddings. Darker cell colors indicate higher representational similarity.}
    \label{fig:geometry_rsa}
\end{figure}

\paragraph{Results.}
We observe a similar pattern as for the CKA analysis on sentence embeddings. \Cref{fig:geometry_rsa} shows that token-level relations remain similar when comparing the code-switching variant to the baseline for all \texttt{eng--nld} models. This indicates that training on code-switched data causes only limited reorganization of the embedding geometry. In contrast, we observe a substantially larger divergence for the \texttt{eng--zho} models, suggesting that code-switching has a much stronger effect for this language pair that does not share lexical overlap. However, different code-switching variants remain highly similar to one another, indicating that they converge toward a shared lexical geometry despite differences in training strategies. 

Overall, training with code-switched data produces a more pronounced representational shift for \texttt{eng--zho} than for \texttt{eng--nld}, on both global structure and relational alignment.

\subsection{Cross-Lingual Transfer}
We now examine how the observed changes in cross-lingual alignment induced by training on code-switched data affect the performance on tasks requiring bilingual knowledge. 
\paragraph{Translation Retrieval.} We test whether the increased alignment on translation pairs suffices to effectively distinguish them from unrelated sentences. We use a similar procedure as \citet{lample2018word} and rank 1,000 candidate sentence representations according to their cosine similarity to the representation of the source sentence.\footnote{Sentences are represented by the average of their token embeddings.} The translation retrieval performance is reported using Precision@$1$ (P@$1$), which measures the percentage of the correct translation that appears among the top-$1$ retrieved sentences. Higher P@$1$ indicates stronger cross-lingual alignment.

\paragraph{Results.}
We report retrieval performance as P@$1$ over all candidate sentences in \Cref{tab:sentence_retrieval_p1} and see a clear difference between the two language pairs. For \texttt{eng--nld}, training with code-switched data improves translation retrieval over the bilingual baseline. This indicates that code-switching increases not only the absolute similarity of translation pairs as shown in \Cref{fig:alignment}, but also their relative distance to other candidate sentences. In contrast, the \texttt{eng--zho} results show a weaker retrieval pattern. All code-switching variants further decrease the already low retrieval precision. While retrieval success remains above the chance level, it indicates that code-switching does not reliably position Chinese sentences closer to their English translations relative to other candidate sentences.

\begin{table}[ht]
    \centering
    \small
    \begin{tabular}{llcc}
        \toprule
        Language pair & Model & \multicolumn{1}{c}{eng $\rightarrow$ X} & \multicolumn{1}{c}{X $\rightarrow$ eng} \\
        \midrule
        \multirow{5}{*}{eng--nld}
        & Baseline    & 13.9 & 14.9 \\
        & Static      & 30.5 & 32.0 \\
        & Curr.       & 20.2 & 22.1 \\
        & POS-Static  & 21.5 & 22.4 \\
        & POS-Curr.   & 17.2 & 19.4 \\
        \midrule
        \multirow{5}{*}{eng--zho}
        & Baseline    & \phantom{0}5.6 & \phantom{0}6.9 \\
        & Static      & \phantom{0}2.7 & \phantom{0}3.0 \\
        & Curr.       & \phantom{0}2.7 & \phantom{0}3.1 \\
        & POS-Static  & \phantom{0}2.8 & \phantom{0}3.1 \\
        & POS-Curr.   & \phantom{0}2.6 & \phantom{0}2.9 \\
        \bottomrule
    \end{tabular}
    \caption{Sentence-level translation retrieval measured by P@1 (\%), showing the percentage of the correct translation that appears among the top-$1$ retrieved sentences. The results are averaged across three random seeds. The P@1 for random retrieval is 0.1\%.}
    \label{tab:sentence_retrieval_p1}
\end{table}

\section{Discussion}
Our complementary analysis of performance aspects and representational structure provides a coherent picture of how training with code-switched data impacts bilingual language models. Our contrasting findings for \texttt{eng--nld} and \texttt{eng--zho} models suggest that the benefit of intra-sentence code-switching depends on how easily the two languages can be integrated within the same training context. For the typologically close \texttt{eng--nld} pair, the bilingual baseline already leads to a reasonably aligned representation space, and code-switching further improves both the accuracy on linguistic minimal pairs and downstream performance. The strongest results are obtained with the unconstrained Static variant. This may indicate that further constraining the \textit{switch point} using linguistic constraints is not necessary if the surface structure of the two languages is highly similar (as is the case for Dutch and English). These findings are compatible with Poplack's (\citeyear{poplack_sometimes_1980}) Equivalence Constraint, under which switching is more likely at points where the surface structures of the two languages can be combined without violating either grammar. 

For the \texttt{eng--zho} models, training with code-switched data fails to improve meaningful bilingual alignment and also leads to weaker grammatical knowledge in Chinese. At the same time, some code-switched models produce higher absolute similarity between \texttt{eng--zho} translation pairs than the bilingual baseline. This suggests that the problem is not simply that the two languages become less similar in the representation space. Rather, the increase in similarity does not translate into better relative alignment, as non-translation pairs may become similarly close. The CKA and RSA results provide further evidence that code-switching substantially reorganizes the Chinese representation space while leaving the English space comparatively stable. These findings suggest that intra-sentence code-switching may alter the organization of Chinese representations without producing the translation-relevant structure needed for effective cross-lingual transfer.

The difficulty of the \texttt{eng--zho} setting may partly arise from the substantially different linguistic and orthographic properties of the two languages, which provide fewer straightforward \emph{switch points} for intra-sentence integration. However, this interpretation should be considered alongside the training constraints of the BabyLM challenge \citep{choshen2026babylmturns4goes}. The limited token budget requires the POS tags and bilingual dictionary to be self-trained from relatively little data, making the resources used to identify and construct switching points relatively noisy. Such noise may have a greater impact when the languages are more distant, where reliable lexical and structural correspondences are already harder to establish. Thus, the observed difference between \texttt{eng--nld} and \texttt{eng--zho} may reflect an interaction between properties of the language pair and the quality of the resources available for code-switching, rather than typological distance alone.

\section{Conclusion}
We tested four cognitively-inspired code-switching variants and examined their effect on bilingual language models for two language pairs. Our results show that training with code-switched data can improve cross-lingual transfer and representational alignment, but its effectiveness seems to depend on the compatibility of the languages. 

For the closely related \texttt{eng--nld} pair, training with code-switched data generally improves linguistic proficiency, downstream performance, and internal cross-lingual alignment compared to the bilingual baseline. The effects of manipulating the \emph{switch rate} and the\emph{switch point} are relatively small, with the Static variant showing slightly better overall performance. This suggests that the introduction of code-switching itself is more influential than the precise choice of when or how frequently language switching occurs for \texttt{eng--nld}. The results for \texttt{eng--zho}, however, show that these benefits do not necessarily generalize to all languages. Code-switching variants perform worse than the bilingual baseline on sentence-level translation retrieval and on grammatical tasks, especially for Chinese. Within the \texttt{eng--zho} code-switching variants, POS-Static tends to perform marginally better than the ones with random \emph{switch point} and the ones with high \emph{switch rate}, suggesting that constraining the switching location and reducing switched content at early training stages may better preserve grammatical structure when code-switching between more distant languages. 

Overall, our findings suggest that training with code-switched data is a promising mechanism for improving cross-lingual transfer, particularly for languages that have high lexical overlap. Future work could therefore investigate whether typological distance remains a significant factor when larger token budgets, higher-quality POS annotations, and more reliable bilingual dictionaries are available.

\section*{Limitations}
Although training with code-switched data appears to support cross-lingual transfer, particularly for \texttt{eng--nld} even under the limited-token constraints, several limitations remain regarding the cognitive plausibility of these models.

Importantly, our approach does not strictly mirror the cognitive and interactional realities of bilingual child language acquisition. In naturalistic settings, parent-child code-switching is highly contextualized and communicatively driven; for instance, bilingual parents frequently utilize one language to instruct an infant about a novel object or concept associated with another language \citep{garcia-1980-function, BAIL_MORINI_NEWMAN_2015}. Our word- and phrase-level substitution framework lacks this semantic and situational grounding, as it relies on a blind stochastic process rather than communicative intent. In addition, the code-switched data we use in this study is synthetic. While we try to introduce high-quality code-switched context, the data are inherently noisy and occasionally invalid, as they are generated via a probabilistic pipeline applied to translated parallel corpora. Another limitation in cognitive plausibility stems from the modeling design: our current pipeline exposes the model to a fixed architectural design from the onset of training, while in human language acquisition, children dynamically expand their vocabulary size, lexical processing speed, and syntactic complexity over time. Incorporating a cognitively plausible architecture remains an important direction for future work.

\section*{Acknowledgments}
We thank the reviewers for the thoughtful and detailed feedback. We thank Irantzu Elcoro Alberdi for the insightful discussion of the literature on code-switching. This research is partially supported by the zukunft.niedersachsen program of the VolkswagenStiftung (L.B., Z.H., L.P.) and by a VENI grant (Vl.Veni.211C.039) from the Nederlandse Organisatie voor Wetenschappelijk Onderzoek (NWO) (L.B.). 

\section*{Author Contributions}
\textbf{Z.H.:} Conceptualization, Analysis, Literature review, Visualization, Writing---original draft, Writing---review \& editing.
\textbf{L.P.:} Conceptualization, Literature review, Writing---original draft (parts of Introduction and Related Work). 
\textbf{L.B.:} Supervision, Conceptualization, Analysis, Writing---review \& editing.

\bibliography{anthology-1,anthology-2,custom}

\newpage

\appendix

\section{Building the Bilingual Dictionaries}
\label{sec:bilingual_dict}
The bilingual dictionaries utilized in our code-switching pipeline are built from parallel subtitle corpora from OpenSubtitles \cite{lison-tiedemann-2016-opensubtitles2016}, obtained via the OPUS platform \cite{tiedemann2012opus}. Prior to statistical alignment, all sentence pairs were filtered to enforce a maximum length constraint of 80 tokens per side and a length ratio of $\le$ 9:1. The specific dataset parameters, release configurations, and resulting sentence pair counts utilized during the alignment process are detailed in Table~\ref{tab:align_data}.

\begin{table}[h]
\centering
\small
\label{tab:corpus_configs}
\begin{tabular}{ccc}
\toprule
\textbf{Languages} & \textbf{Release} & \textbf{Aligned Sentence Pairs} \\
\midrule
\texttt{eng--nld} & v2018 & 325,874 \\
\texttt{eng--zho} & v2024 & 504,020 \\
\bottomrule
\end{tabular}
\caption{Parallel Corpus Configurations for Bilingual Lexicon Extraction}
\label{tab:align_data}
\end{table}

To learn word-level translation mappings, we perform unsupervised word alignment using MGIZA \citep{gao-vogel-2008-parallel}, a multi-threaded extension of GIZA++ \citep{och03:asc}. The preprocessing pipeline first generates internal source vocabularies and sentence co-occurrence structures via \texttt{snt2cooc}, while \texttt{mkcls} \citep{och-1999-efficient} clusters words into statistical frequency classes. Word alignments are trained using the Expectation--Maximization (EM) algorithm \citep{brown-etal-1993-mathematics}. To bootstrap baseline lexical translation probabilities, the optimization routine begins with 5 iterations of IBM Model 1. The alignment pipeline then transitions to 3 iterations each of IBM Models 3 and 4 to account for word-order reordering (distortion) and one-to-many word mappings (fertility). Models are trained symmetrically in both directions ($\text{source} \rightarrow \text{target}$ and $\text{target} \rightarrow \text{source}$). Finally, the resulting directional alignments are merged using the standard \texttt{grow-diag-final-and} symmetrization heuristic \citep{koehn-etal-2003-statistical} to extract high-confidence word and multi-word alignment links.

\paragraph{English--Dutch Lexicon Extraction}
Prior to statistical alignment, the English and Dutch text is standardized by lowercasing and applying whitespace tokenization. Once the unified bidirectional alignments are produced by the Pharaoh-formatted alignment map, a unigram extraction layer processes the aligned sentences to count every valid one-to-one mapping across the corpus. To eliminate structural alignment noise, spurious translations, and low-frequency accidental links, we enforce a strict minimum co-occurrence threshold where a word translation pair must appear $\geq 5$ times in the corpus to be retained. Punctuation marks are automatically stripped, and the final filtered lexicon maps individual source unigrams directly to their most probable target counterparts.

\paragraph{English--Chinese Lexicon Extraction}
Prior to running the alignment models, the text is processed without relying on any pretrained segmenters or tokenizers: the English data is lowercased and whitespace-tokenized, whereas the Chinese sentences are split strictly at the character level, treating each individual character as an autonomous token. From the unified bidirectional alignments, the lexicon is constructed via a dual-layer extraction architecture to properly handle structural differences:
\begin{enumerate}
    \item \textbf{Unigram Layer:} Captures and counts every direct one-to-one alignment link across the corpus (e.g., mapping individual English words to single Chinese characters).
    \item \textbf{Phrase Layer:} Extracted via standard Moses-style phrase extraction to recover compound multi-character concepts. For each alignment anchor, the algorithm identifies all contiguous English and Chinese spans (up to a maximum boundary constraint of 5 English tokens and 6 Chinese characters), where every internal alignment link is strictly consistent, and no links cross outside the candidate span boundaries. 
\end{enumerate}
Once multi-word English expressions and multi-character Chinese compounds are recovered, contiguous Chinese characters within an extracted phrase are joined into a unified, non-whitespace string (e.g., fusing character-level outputs to form a multi-character word). The unigram and phrase layers are subsequently merged into a master dictionary. Mappings are filtered out if they have fewer than a minimum co-occurrence frequency threshold ($\geq 2$ occurrences), punctuation marks are stripped, and English-only function-word phrases are discarded. Finally, a subsumption pruning pass removes shorter dictionary entries if they are strictly dominated by a longer phrase that possesses an equal or higher co-occurrence frequency. 

\paragraph{Runtime Dictionary Utilization}
Each generated entry in both lexicons tracks total occurrence counts alongside the empirical translation probability $P(\text{target} \mid \text{source})$, representing how frequently a specific translation appears among all candidate options for a given source term. At runtime, the data pipeline queries these ranked candidates directly, choosing the highest-probability option.

\section{POS tagger Trainer} 
\label{sec:pos_tagger}
In this work, we train part-of-speech (POS) taggers for English, Chinese, and Dutch using the Universal Dependencies treebanks \citep{nivre-etal-2017-universal}. Each dataset is provided in CoNLL-U format and is split into standard training, development, and test partitions. 

We employ a BiLSTM-CRF architecture implemented in PyTorch. The model is trained from scratch without the use of pretrained word embeddings. Instead, it learns both word-level and character-level representations directly from the training data. Each token is represented as the concatenation of a learned word embedding and a character-level representation obtained via a convolutional neural network over characters. This allows the model to better handle rare and out-of-vocabulary words. The sequence of token representations is then processed by a bidirectional LSTM, which captures contextual information from both the left and right contexts of each word. The resulting contextualized representations are passed through a linear layer to produce emission scores over POS tags. A conditional random field (CRF) layer is used on top to model dependencies between adjacent tags and to decode the globally optimal tag sequence.

For data preparation, sentences are extracted from the CoNLL-U files as sequences of tokens paired with their gold POS tags. We construct vocabularies for words, characters, and POS tags based on the training set. During inference, unseen words are mapped to a special \texttt{UNK} token.

The model is trained using the Adam optimizer with a learning rate of $10^{-3}$, a batch size of 32, and gradient clipping to improve training stability. The training objective is the negative log-likelihood of the correct tag sequence under the CRF layer. After each epoch, performance is evaluated on the development set using token-level accuracy. The best-performing model on the development set is saved separately for each language. Early stopping is applied if development accuracy does not improve for three consecutive epochs.

At evaluation time, the saved model checkpoints are used to compute performance on both the development and test sets. We report overall token-level accuracy as well as per-tag accuracy for major POS categories. The models achieve approximately 82\% accuracy on English, 85\% on Chinese, and 83\% on Dutch test sets. Performance is generally strongest on frequent and syntactically regular tags, while lower performance is observed on rare categories such as interjections and foreign words.

Once trained, the models assign a POS tag to each token in an input sentence. For example, in English, \textit{orange} may be labeled as an ADJ and \textit{cat} as a NOUN. However, in Chinese, as a word can be formed by several characters, the POS tagger can assign one tag to a compound word. For example, \zh{橘猫} (orange cat) can be labeled as NOUN.

\section{Data Licensing}
The training data used for our models follow the licensing conditions described by \citet{jumelet-etal-2026-babybabellm}, who report that, during their data collection effort, they verified that all data were released with licenses permitting academic research, such as Creative Commons licenses or Public Domain.

For the construction of the bilingual dictionaries, we used the OpenSubtitles2016 corpus \citep{lison-tiedemann-2016-opensubtitles2016}, which is distributed through the OPUS collection and was made freely available to the research community. For training the POS tagger, we used the Universal Dependencies (UD) English Web Treebank (EWT) \citep{silveira-etal-2014-gold}, Dutch LassySmall \citep{bouma-van-noord-2017-increasing}, and Chinese PUD treebanks \citep{mcdonald-etal-2013-universal}. The English-EWT and Dutch-LassySmall treebanks are licensed under CC BY-SA 4.0, while the Chinese-PUD treebank is licensed under CC BY-SA 3.0. We used these resources in accordance with their respective licensing terms.

Finally, the Dutch--English cognate dataset used for cross-lingual alignment is released under the Creative Commons Attribution--NonCommercial (CC BY-NC) license, which permits sharing and adaptation with appropriate attribution but prohibits commercial use.

\begin{table*}[tb]
\centering
\small
    \begin{tabular}{p{4.2cm}p{8.8cm}}
    \hline
    \textbf{Parameter} & \textbf{Value} \\
    \hline
    Model architecture&
    12-layer GPT-2 with hidden size 768 and 12 attention heads. \\
    Tokenizer &
    16,384-token shared bilingual BPE vocabulary. \\
    Optimizer &
    AdamW. \\
    Learning Rate &
    $5\times10^{-5}$. Cosine decay with 1\% linear warmup. \\
    Batch size &
    16. \\
    Sequence length &
    512 tokens. \\
    Randomization &
    Independent random seeds (13, 42, 123) for model initialization and code-switch decisions. \\
    \hline
    \end{tabular}
\caption{Training hyperparameters and configuration.}
\label{tab:training}
\end{table*}
\section{Hyperparameter}
The hyperparameters of model training are described in \Cref{tab:training}.

\section{Cognates Alignment}
\begin{figure}[th]
    \centering
    \includegraphics[width=\columnwidth]{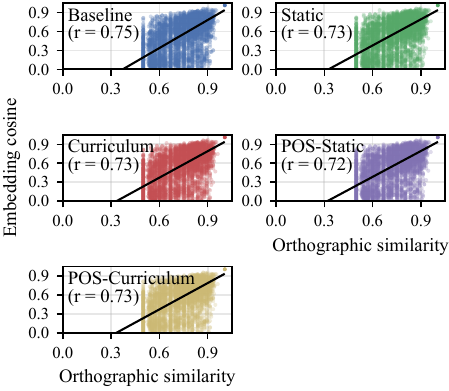}
    \caption{Relationship between orthographic similarity and word embedding cosine similarity for English--Dutch cognates ($n = 5,285$ pairs). Lower Pearson correlation indicates better cross-lingual alignment, especially for cognates with less similar surface forms.}
    \label{fig:cognates_all_models}
\end{figure}

\Cref{fig:cognates_all_models} shows the Pearson correlation between orthographic similarity and word embedding cosine similarity for English--Dutch cognates. The figure shows the results for the bilingual baseline and all four code-switching variants.

\label{app:cognates_alignment}

\newpage

\section{BabyLM Benchmark}
\Cref{tab:zero-shot_tasks} and \Cref{tab:fine-tune_tasks} list all the tasks being evaluated for \texttt{eng-nld} and \texttt{eng-zho} models.

\label{sec:task_list}
\begin{table*}[h]
\centering
\small
\begin{tabular}{p{3.2cm}p{4.2cm}p{5.8cm}}
\hline
\textbf{Evaluation Dimension} & \textbf{eng--nld Tasks} & \textbf{eng--zho Tasks} \\
\hline

Grammatical competence &
\texttt{blimp\_babylm\_filtered}$^{1}$, 
\texttt{multiblimp\_eng}$^{2}$,
\texttt{blimp\_nl}$^{3}$,
\texttt{multiblimp\_nld}$^{2}$ &
\texttt{blimp\_babylm\_filtered}$^{1}$,
\texttt{multiblimp\_eng}$^{2}$,
\texttt{zhoblimp}$^{4}$ \\

Commonsense reasoning &
\texttt{hellaswag\_en\_mubench}$^{5}$,
\texttt{winogrande\_en\_mubench}$^{5}$,
\texttt{hellaswag\_nl\_mubench}$^{5}$,
\texttt{winogrande\_nl\_mubench}$^{5}$ &
\texttt{hellaswag\_en\_mubench}$^{5}$,
\texttt{winogrande\_en\_mubench}$^{5}$,
\texttt{hellaswag\_zh\_mubench}$^{5}$,
\texttt{winogrande\_zh\_mubench}$^{5}$ \\

Discourse understanding &
\texttt{xstorycloze\_en\_mubench}$^{5}$,
\texttt{xstorycloze\_nl\_mubench}$^{5}$ &
\texttt{xstorycloze\_en\_mubench}$^{5}$,
\texttt{xstorycloze\_zh\_mubench}$^{5}$ \\

Compositional semantics &
\texttt{xcomps\_nl}$^{6}$ &
\texttt{xcomps\_zh}$^{6}$ \\

\hline
\end{tabular}
\caption{Zero-shot evaluation benchmarks grouped by evaluation dimension. $^{1}$\cite{choshen2026babylmturns4goes};
$^{2}$\cite{jumelet-etal-2026-multiblimp};
$^{3}$\cite{suijkerbuijk-etal-2025-blimp};
$^{4}$\cite{liu2025systematicassessmentlanguagemodels};
$^{5}$\cite{han-etal-2026-mubench};
$^{6}$\cite{he-etal-2025-xcomps};
}
\label{tab:zero-shot_tasks}
\end{table*}

\begin{table*}[h]
\centering
\small
\begin{tabular}{p{5.8cm}p{4.2cm}p{3.2cm}}
\hline
\textbf{Evaluation Dimension} & \textbf{eng--nld Tasks} & \textbf{eng--zho Tasks} \\
\hline

Science and general reasoning &
\texttt{arc\_mubench}$^{1}$ &
\texttt{arc\_mubench}$^{1}$ \\

Reading comprehension &
\texttt{belebele}$^{2}$ &
\texttt{belebele}$^{2}$ \\

Factual knowledge &
\texttt{bmlama\_mubench}$^{1}$ &
\texttt{bmlama\_mubench}$^{1}$ \\

Natural language inference &
\texttt{mnli\_mubench}$^{1}$, \texttt{xnli} (eng)$^{3}$,
\texttt{mnli\_mubench} (nld)$^{1}$ &
\texttt{mnli\_mubench}$^{1}$, \texttt{xnli}$^{3}$ \\

Topic classification &
\texttt{sib200}$^{4}$ &
\texttt{sib200}$^{4}$ \\

Truthfulness and calibration &
\texttt{truthfulqa\_mubench}$^{1}$ &
\texttt{truthfulqa\_mubench}$^{1}$ \\

Language-specific understanding &
\texttt{include} (nld)$^{5}$ &
\texttt{include} (zho)$^{5}$ \\

\hline
\end{tabular}
\caption{Fine-tuning evaluation benchmarks grouped by evaluation dimension. $^{1}$\cite{han-etal-2026-mubench};
$^{2}$\cite{bandarkar-etal-2024-belebele};
$^{3}$\cite{conneau-etal-2018-xnli};
$^{4}$\cite{adelani-etal-2024-sib};
$^{5}$\cite{romanou2024include};}
\label{tab:fine-tune_tasks}
\end{table*}

\newpage

\section{Per-Category Results}
\label{app:detailed_results}
\Cref{tab:en_nl_results} and \Cref{tab:en_zh_results} present the results of all models across each task.

\begin{table*}[ht]
\centering
\small
\begin{tabular}{llccccc}
\toprule
\textbf{Task} & \textbf{Setting} & \textbf{Baseline} & \textbf{Static} & \textbf{Curriculum} & \textbf{POS-Static} & \textbf{POS-Curriculum} \\
\midrule
\multicolumn{7}{l}{\textit{Zero-Shot}} \\
\midrule
BLiMP & eng & 72.6 & \textbf{73.1} & 72.6 & 72.7 & 72.4 \\
MultiBLiMP & eng & 87.7 & \textbf{88.8} & 87.9 & 88.1 & 88.3 \\
Hellaswag & eng & 26.5 & 26.5 & 26.5 & 26.4 & \textbf{26.6} \\
Winogrande & eng & 50.8 & 50.3 & 50.3 & \textbf{51.1} & 50.1 \\
XStoryCloze & eng & 49.1 & 49.0 & 48.6 & \textbf{49.2} & 48.6 \\
PIQA parallel & eng & 22.8 & 20.4 & 20.1 & \textbf{23.3} & 19.7 \\
PIQA nonparallel & eng & 49.5 & 51.0 & \textbf{51.3} & 50.7 & 50.0 \\
BLiMP & nld & 80.8 & 80.9 & 80.6 & \textbf{81.1} & 80.1 \\
MultiBLiMP & nld & 93.8 & \textbf{94.0} & 93.5 & 93.5 & 93.1 \\
Hellaswag & nld & 26.5 & 26.5 & \textbf{26.6} & 26.3 & \textbf{26.6} \\
Winogrande & nld & 49.8 & \textbf{50.2} & 49.6 & 49.8 & \textbf{50.2} \\
XCOMPS & nld & 53.8 & 54.0 & \textbf{54.3} & 53.9 & \textbf{54.3} \\
XStoryCloze & nld & 48.4 & \textbf{48.6} & 47.7 & 48.2 & 48.3 \\
PIQA parallel & nld & 22.7 & 22.3 & 22.3 & 18.8 & \textbf{23.6} \\
PIQA nonparallel & nld & 55.0 & 55.7 & 54.7 & \textbf{56.7} & 50.7 \\
\midrule
\multicolumn{7}{l}{\textit{Fine-Tuned}} \\
\midrule
ARC & eng & \textbf{25.6} & 24.7 & 25.1 & 25.1 & 24.4 \\
ARC & nld & \textbf{24.9} & 23.8 & \textbf{24.9} & \textbf{24.9} & 24.1 \\
Belebele & eng & 25.0 & 23.9 & 23.9 & 23.9 & \textbf{25.6} \\
Belebele & nld & 23.9 & \textbf{28.2} & 25.0 & 25.0 & 28.1 \\
BM-LAMA & eng & 9.7 & 10.2 & 11.3 & 9.7 & \textbf{11.8} \\
BM-LAMA & nld & 11.5 & \textbf{12.0} & 10.3 & 10.2 & 10.0 \\
MNLI & eng & \textbf{50.3} & \textbf{50.3} & 48.5 & 49.2 & 49.7 \\
MNLI & nld & 49.6 & 49.8 & 48.9 & 51.0 & \textbf{51.3} \\
SIB-200 & eng & 59.8 & \textbf{78.3} & 76.2 & 58.8 & 77.7 \\
SIB-200 & nld & 39.8 & 72.2 & 55.8 & 56.7 & \textbf{72.7} \\
TruthfulQA & eng & \textbf{27.7} & \textbf{27.7} & 24.7 & 25.6 & 25.4 \\
TruthfulQA & nld & 26.2 & 22.3 & 23.2 & \textbf{27.1} & 25.0 \\
XNLI & eng & 47.3 & 45.2 & 45.9 & 47.2 & \textbf{47.4} \\
INCLUDE & nld & 29.5 & \textbf{33.6} & 28.6 & \textbf{33.6} & 27.2 \\
\bottomrule
\end{tabular}
\caption{Evaluation results for \texttt{eng--nld}. The results are averaged across three random seeds.}
\label{tab:en_nl_results}
\end{table*}

\begin{table*}[ht]
\centering
\small
\begin{tabular}{llccccc}
\toprule
\textbf{Task} & \textbf{Setting} & \textbf{Baseline} & \textbf{Static} & \textbf{Curriculum} & \textbf{POS-Static} & \textbf{POS-Curriculum} \\
\midrule
\multicolumn{7}{l}{\textit{Zero-Shot}} \\
\midrule
BLiMP filtered & eng & 71.3 & \textbf{71.8} & 71.7 & 70.0 & 71.5 \\
MultiBLiMP & eng & 86.0 & \textbf{87.2} & 86.9 & 85.2 & 86.5 \\
Hellaswag & eng & \textbf{26.5} & \textbf{26.5} & 26.4 & 26.2 & 26.4 \\
Winogrande & eng & 50.9 & 50.6 & 50.8 & 48.9 & \textbf{51.1} \\
XStoryCloze & eng & \textbf{50.2} & 48.8 & 49.0 & 47.5 & 48.6 \\
PIQA parallel & eng & 25.6 & 25.9 & 27.2 & \textbf{28.8} & 26.9 \\
PIQA nonparallel & eng & \textbf{52.3} & 49.3 & 50.0 & 51.3 & 48.0 \\
Hellaswag & zho & \textbf{26.9} & 25.9 & 25.9 & 25.8 & 26.0 \\
Winogrande & zho & 49.3 & 51.0 & 50.6 & 50.8 & \textbf{51.1} \\
XCOMPS & zho & \textbf{53.4} & 50.3 & 50.3 & 50.8 & 50.7 \\
XStoryCloze & zho & \textbf{48.8} & 46.2 & 46.3 & 46.6 & 46.6 \\
ZhoBLiMP & zho & \textbf{76.0} & 54.9 & 55.1 & 54.5 & 56.0 \\
Hanzi structure & zho & \textbf{56.2} & 50.1 & 50.6 & 50.6 & 50.9 \\
Hanzi pinyin & zho & 49.2 & 50.9 & \textbf{51.4} & 50.0 & 51.0 \\
PIQA parallel & zho & \textbf{21.8} & 19.7 & 18.8 & 20.7 & 17.5 \\
PIQA nonparallel & zho & \textbf{50.5} & 46.0 & 47.3 & 49.3 & 48.3 \\
\midrule
\multicolumn{7}{l}{\textit{Fine-Tuned}} \\
\midrule
ARC & eng & 24.8 & 23.8 & \textbf{26.2} & 24.4 & 24.5 \\
ARC & zho & 24.6 & 24.2 & \textbf{25.5} & 25.1 & 25.1 \\
Belebele & eng & 25.0 & 25.9 & 25.0 & \textbf{26.1} & \textbf{26.1} \\
Belebele & zho & 25.0 & 22.5 & 24.2 & 23.9 & \textbf{26.9} \\
BM-LAMA & eng & 10.1 & 10.3 & 10.9 & 10.2 & \textbf{11.9} \\
BM-LAMA & zho & 9.5 & 10.2 & 10.1 & \textbf{10.9} & 10.5 \\
MNLI & eng & 46.4 & 49.8 & \textbf{56.0} & 49.4 & 50.3 \\
MNLI & zho & \textbf{45.8} & 38.8 & 42.2 & 39.1 & 41.9 \\
SIB-200 & eng & 36.8 & \textbf{76.8} & 76.2 & 56.8 & 74.7 \\
SIB-200 & zho & 40.0 & \textbf{51.3} & 29.3 & 37.0 & 36.8 \\
TruthfulQA & eng & 21.7 & \textbf{26.5} & 24.4 & 23.2 & 25.0 \\
TruthfulQA & zho & 22.3 & 25.3 & \textbf{27.7} & \textbf{27.7} & 23.5 \\
XNLI & eng & 45.2 & 45.6 & \textbf{47.7} & 44.6 & 43.3 \\
XNLI & zho & 45.1 & \textbf{46.1} & 45.6 & 45.4 & 45.1 \\
INCLUDE & zho & 26.8 & 27.1 & 26.5 & 22.0 & \textbf{28.3} \\
\bottomrule
\end{tabular}
\caption{Evaluation results for \texttt{eng--zho}. The results are averaged across three random seeds.}
\label{tab:en_zh_results}
\end{table*}

\end{document}